# Adaptive Road Configurations for Improved Autonomous Vehicle-Pedestrian Interactions using Reinforcement Learning

Qiming Ye*, Yuxiang Feng, *Member, IEEE*, Jose Javier Escribano Macias, Marc Stettler, Panagiotis Angeloudis

*Abstract*—The deployment of Autonomous Vehicles (AVs) poses considerable challenges and unique opportunities for the design and management of future urban road infrastructure. In light of this disruptive transformation, the Right-Of-Way (ROW) composition of road space has the potential to be renewed. Design approaches and intelligent control models have been proposed to address this problem, but we lack an operational framework that can dynamically generate ROW plans for AVs and pedestrians in response to real-time demand. Based on microscopic traffic simulation, this study explores Reinforcement Learning (RL) methods for evolving ROW compositions. We implement a centralised paradigm and a distributive learning paradigm to separately perform the dynamic control on several road network configurations. Experimental results indicate that the algorithms have the potential to improve traffic flow efficiency and allocate more space for pedestrians. Furthermore, the distributive learning algorithm outperforms its centralised counterpart regarding computational cost (49.55%), benchmark rewards (25.35%), best cumulative rewards (24.58%), optimal actions (13.49%) and rate of convergence. This novel road management technique could potentially contribute to the flow-adaptive and active mobility-friendly streets in the AVs era.

*Index Terms*—Autonomous Vehicles, Pedestrians, Smart City, Intelligent Transport System, Reinforcement Learning, Infrastructure management

## I. INTRODUCTION

Autonomous vehicles (AVs) are predicted to be the next prevailing mode of urban mobility [1]. According to the SAE International standards [2], AVs classified under the high automation (Level 4) and full automation (Level 5) categories would take over most or all (respectively) driving-related duties, with the vehicle systems performing the functions of environment perception, localisation, mapping, path planning and driving control [3], [4].

As the related technologies continue maturing, AVs are expected to significantly improve the safety and efficiency of urban transport flows, eliminating potential human errors in driving. For example, recent studies imply that AVs with Level 4 Automation have the potential to prevent 28% of the fatal pedestrian-car crashes, while Level 5 Automation would prevent up to 73% of such severe accidents [5]. It is predicted that Level 4 AVs can be realised by 2030, and their market share may reach 50% before 2045 [6], [7].

All authors are with the Centre for Transport Studies, Department of Civil and Environmental Engineering, Imperial College London, South Kensington, SW7 2AZ, London.
* Corresponding author: Qiming Ye (qiming.ye18@imperial.ac.uk)

As a result of their deployment, platoon-based driving would be possible, which reduces AV headways and reaction times drastically [4], thereby decreasing the number of lanes required for road traffic. The results of recent studies on this topic [8], [9] suggest that AVs can drive on a four-lane road with the same efficiency as human-driven vehicles on a six-lane road.

Recent studies have indicated that city centres could experience a rise in AV-based trips of approximately 25% during rush hours, with a corresponding increase in travel costs of 5.5% [10]. Consequently, the mean volume-to-capacity ratio (V/C) is estimated to rise by 7.99% and 8.44% for respective peaks [1]. Compared to outskirts, travel costs at city centres would drop by 12.1% [10]. Additionally, the total travel time is likely to decrease by 20%~25% during off-peak hours [11], which will have a substantial positive impact on V/C conditions.

Flexible Right-of-Way (ROW) strategies would be preferable to static ROW configurations that restrict users to specific parts of the road space [12], [13], allowing the layout to be adapted to varying traffic demand patterns at different locations and times of a day. It might be possible to de-motorise some lanes during off-peak hours or in peripheral regions to encourage active mobility and local street activities [14], [15]. During rush hours on major thoroughfares in city centres, on the other hand, it is vital to remain the capacity of delivering basic Levels Of Service (LOS) for vehicular traffic, and ensure that all road users have convenient access.

Prior works on AV-aware adaptive streets pursued the adoption of design perspectives [8], [14], [16] or novel modelling techniques [17]–[19]. Despite the advances these studies achieved so far, the key limitations are the lack of operational and demand-responsive ROW management schemes across the entire spectrum of road users. As such, we consider the absence of such a dynamic approach to ROW compositions as a research gap. It is expected that ROW compositions could evolve to respond to changing demand patterns of road users and their potential interactions.

The research problem is defined as a multi-agent based stochastic control problem involving a time-discrete finite-horizon transport system. To solve it, we develop a microscopic traffic simulation-integrated Reinforcement Learning (RL) model. The model aims to achieve a balance between traffic flow efficiency and expanding space for non-vehicular road users by continuously evolving the control policy that determines sequential ROW compositions at regular intervals (e.g. 30 minutes).

For the purposes of this study, we will be using a Deep Deterministic Policy Gradient (DDPG) algorithm [20]. The key requirement of this modelling framework is a model-free RL approach due to the fact that the system dynamics are unknown. Additionally, an off-policy learning mechanism is employed to ensure sample efficiency and flexibility when exploring the action space [21]. In this article, we propose a Multi-Agent DDPG (MADDPG) algorithm as an innovative learning approach compared with its centralised counterpart, DDPG. They are different in terms of learning paradigms, reward distributions, Actor-Critic (AC) network architecture, and ways of replaying interaction experiences.

The principal contributions are outlined as follows:

1) In our framework, a model-free off-policy RL algorithm is combined with microscopic traffic simulations to continuously evolve the ROW compositions of road space.
2) As the proposed method is adaptable, it can be applied to a variety of traffic scenarios, including fully AVs, pure human traffic, and mixed traffic at diverse levels.
3) Two learning paradigms are embedded in this framework, and their learning performances are compared based on computational cost, benchmark reward, explored optima, and convergence speed.
4) This study analyses the impact of noise disturbance on learning performances in order to identify effective strategies that balance exploration and exploitation in the high-dimensional space of action.

As far as we know, this study is the first attempt to use a RL method to solve ROW control problems. The paper offers a benchmark as well as a novel operational approach to planning and managing AV traffic-adaptive streets.

The remainder of this paper is organised as follows: Section II provides the background of the research. Section III formulates the ROW control problem and presents the structure of our RL model. Experiment set-up and results are explained in Sections IV and V. Finally, Section VI concludes this study and recommends future research directions.

## II. Background

A summary of previous studies on the design approaches and modelling techniques for managing the future AV traffic-adaptive road space is provided in Table. I. The remainder of this chapter summarises key features of studies that we surveyed, and specifies a set of methodological requirements that we used when designing our modelling framework.

TABLE I
DESIGN AND MANAGEMENT APPROACHES OF SURVEYED LITERATURE

| Road Elements | Design Approaches | Control Models |
|---|---|---|
| lane assignment | [8], [14], [16] | [22], [23] |
| lane reversal control | [8], [14], [16] | [19], [24] |
| sidewalk and crossing | [14]–[16] | [18] |
| cycleway | [14]–[16] | - |
| shared space | [8] | - |
| curb space | [14], [15] | [17] |
| traffic signal | - | [25], [26] |

### A. Design Approaches to AV Streets

The UK Highway Code grants different categories of road users the right to use a publicly owned but highly regulated road space, in accordance with hierarchical orders of priority [12]. This right is referred to as the Right-Of-Way (ROW) and has been physically codified into the form of Complete Streets (CS) in many countries [13], [27]. As a general rule, the CS scheme divides the street surface into the carriageway and the street-side section [28]. There are several functional zones within a street-side section, including a facility belt, a front area, and the sidewalk, with an aggregate width exceeding 1.5*m* [28], [29]. The standard driving lanes are designed to be 3.0*m* to 3.5*m* in width [28].

Almost all present proposals for AV streets adopt the concept of CS in the context of future AV transport [15]. One of their underlying motivations for reshaping streets to accommodate AVs is to re-prioritise the ROW of active mobility [8]. A holistic ROW plan is usually included in their proposals, which comprise AV traffic dedicated lanes, transit lanes, sidewalks, cycle-ways, and flex zones [14], [28]. It reduces traffic and parking spaces with increased street-side areas to meet the demands of non-motorised users [15].

As one example, the National Association of City Transportation Officials (NACTO) recommends several ROW layouts depending on the types of roads [14]. Their design goals are polarised, including wider sidewalks, protected cycle-ways and greenery [8], [28]. Their plans of the carriageway still follow the structure of designing CS [30], as per Eq. 1. [1]

$$l = \lceil \frac{\text{AADT} \times \text{KF} \times \text{DF}}{f_{\text{PH}} \times \text{MSF}_i \times f_{\text{HV}} \times f_p} \rceil \quad (1)$$

Three primary problems arise from using design-based approaches. A major drawback is that they heavily rely on empirical data derived from future scenarios where AVs will be widely deployed, which is not available at the moment. Second, the lack of explicit models makes it difficult to calibrate and replicate their designs. Finally, CS layouts that use fixed schemes instead of demand-responsive operational schemes could pose problems because AV traffic patterns would change constantly.

### B. Intelligent Control Methods for AV Streets

To optimise the utilisation of road space for future AV traffic, several methods have been proposed, including dynamic lane control [18], [19], [22], lane allocation [23], curbside management [17], and Traffic Signal Control (TSC) [25], [26]. In addition, present studies [31]–[33] combined optimisation techniques with microscopic traffic simulation models to solve optimisation or control problems related to specific road infrastructure.

As an example, a multi-agent Q-learning model, developed by Gunarathna et al (2020) [19] proposed a model to control the directions of lanes of a highway section. In the optimised scenario, travel time was reduced by 20% compared with business as usual. In another study [17], the distribution of curbside parking lanes in an urban block were optimised, with traffic delays of AV fleet reduced by 28%.

---

[1] AADT: annual average daily traffic (*veh/h*). *KF*: a certain proportion of AADT occurring in the peak hour (K-factor). DF a certain proportion of the peak direction (D-factor). $f_{\text{PH}}$: the peak hour factor. MSF: certain level of maximum service flow rate. $f_{\text{HV}}$: the heavy vehicle adjustment factor. $f_p$: the familiarity adjustment factor of drivers, respectively.

Similarly, Ye et al (2022) [18] optimised the width of the carriageway to maximise the traffic flow efficiency of an intersection. In 18.2% of all simulated time-steps, a lane-width space can be re-assigned to the street-side section.

These modelling techniques offer the major advantage of treating road infrastructure as a resource that can be controlled in real-time, incorporating with intelligent AV traffic management. This allows us to mitigate the drawbacks of fixed ROW layouts using advanced optimisation techniques. However, the crucial flaw is that their modelling objectives are usually motorised traffic-oriented, which primarily focus on travel costs [19], [24] and traffic delay [17], ignoring diverse demands from active mobility [28]. In summary, current measurements fall far short of achieving operational ROW management while satisfying the needs of both motorised and non-motorised users. In addition, very few studies [18] have tested their proposed schemes under complex typologies of road networks.

Similar to the portfolio management (asset allocation) problem [34], the ROW control problem is a stochastic control problem with nonlinear state transitions and unknown reward mechanisms. The control sequences can be expressed by Markov Decision Processes (MDPs), which comprise successive decisions to make at discrete time steps and additive cost overtime [35]. Meanwhile, the problem presented here is a continuous control problem, which determines respective proportion of road space assigned to each sidewalk in the continuous domain ($B \subseteq R_{(0,1)}$). According to the principle of optimality [36], solving this problem is equivalent to solving all tail sub-problems which consider the expected rewards of future steps.

### C. Convex Optimisation Control Policies and Reinforcement Learning Methods

Convex optimisation control policies (COCPs) are widely applied to solving control problems of explicit system dynamics, namely state transition probabilities and structured reward functions [35]. Through solving a sequence of convex optimisation problems, they find optimal policies to determine sequential decisions [37]. COCPs include Linear Quadratic Regulator (LQR), Quadratic Programs (QP), Model Predictive Control (MPC) and Exact Dynamic Programming (EDP) [35], [37]. They are different from static optimisation methods in that they estimate the potential costs of future steps when evaluating a present decision, whereas static convex optimisation methods only consider that of a single step [37].

Large-scale real-world stochastic control problems present challenges to conventional COCPs. First, the system dynamics are usually nonlinear and unknown [38]. As a result, LQRs, QPs, and MPCs require extensive assumptions, which introduce additional bias to their models [39], [40]. Second, the scalability of COCPs represents a critical concern once the action and state spaces become large and continuous [41]. Further, COCP methods are model-dependent and problem-specific. Thus, generalising them to another problem with a distinctive context can be difficult [42].

Assisted with neural networks and advanced simulation techniques, Reinforcement Learning (RL) methods can solve a wide variety of real-world stochastic control problems which were intractable in the past [20], [21], [41]. Instead of relying on prior knowledge of system dynamics, RLs (e.g. model-free RLs) improve control policies by letting intelligent agents trial-and-error interact with the system [43]. They approximate optimal policies non-exhaustively in order to avoid the curse of dimensionality [40]. Additionally, RLs (e.g. off-policy RLs) have higher degrees of freedom in the control policy, and they may arrive at to good policies despite sample shortage [44].

Similar to COCPs, model-based RLs requires certain levels of understanding of system dynamics [40], whereas model-free RLs learn from direct interactions or past interaction samples [41]. A variety of model-free RLs have been invented, including Q-learning, SARSA, Deep Q-Network (DQN), Actor-Critic (AC), Trust Region Policy Optimisation (TRPO), Policy Gradient (PG), Proximal Policy Optimisation (PPO), Deep Deterministic Policy Gradient (DDPG) algorithms [20], [43]. Besides considering the complexity of system dynamics, selecting the appropriate RL algorithm is influenced by the following factors: (1) simplicity of control policy, (2) continuity of control space, (3) simulation cost, and (4) sample efficiency [41], [43], [44].

Regarding the control policy of the ROW control problem, it simply determines respective proportion of road space assigned to sidewalks. Therefore, directly optimising the control policy rather than conducting value iteration seems more intuitive and feasible. Policy gradient algorithms, such as PG, AC and DDPG, are capable of solving this type of continuous control task. In contrast, value learning algorithms like Q-learning, SARSA, and DQN, which only operate in discrete control space, are not feasible [45]. Furthermore, policy gradient methods can uniform the action space of a multi-agent system through nondimensionalisation [18], whereas value learning methods cannot.

Last but not least, the sample efficiency of an algorithm represents a crucial factor worth considering when simulating a large-scale transport system, and where sampling data is extremely expensive [44]. On-policy algorithms, like PG, TRPO and SARSA, which embed uniform policies for sampling actions and optimising policies, massively rely on data generated from direct interactions. Off-policy algorithms, such as Q-learning, DQN, and DDPG, separate policy training from updating target policies [46]. By doing so, action exploration can be conducted without requiring time-consuming interactions, which greatly alleviates the sample shortage problem, as well as guaranteeing the stability of convergence [44].

Based on the factors outlined above, DDPG [20] provides substantial advantages to solve this ROW control problem compared to other well-known approaches. It is a model-free, policy gradient-based, off-policy RL method that evolves sequential decisions in the continuous action domain. Similar to the DQN, DDPG deploys an experience replay buffer and target AC networks to improve sample efficiency and flexibility in action exploration. The trust region enforcement mechanism introduced by DDPG allows for more stable convergence than PGs, while preserving sufficient freedom in exploring the action space, just as TRPOs and PPOs do [47]. This algorithm has been deployed to multiple sequential-decision problems such as Traffic Signal Control (TSC) [25], [26] and dynamic lane control problems [17], [48].

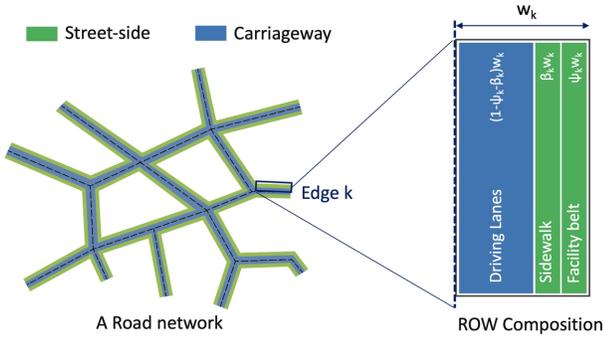

Fig. 1. The Right-Of-Way composition of a road edge inside a network

TABLE II
NOTATIONS OF THE OPTIMAL ROW CONTROL MODEL

| Notations | | Domains |
|---|---|---|
| $K$ | set of edges (agents) | $K \subset N_{[0,+\infty)}$ |
| $T$ | set of time steps | $T \subset N_{[0,+\infty)}$ |
| $N_{k,h}$ | set of operational vehicles on $k$ at $h$ | $N_{k,h} \subset N_{[0,+\infty)}$ |
| $P_{k,h}$ | set of active pedestrians on $k$ at $h$ | $P_{k,h} \subset N_{[0,+\infty)}$ |
| $H_t$ | set of observation at $t$ | $H_t \subset N_{[0,+\infty)}$ |
| $O, D$ | sets of origins and destinations | |
| $L_{k,t}$ | domain of lane numbers | $L_{k,t} \subset N_{[1,+\infty)}$ |
| $B_{k,t}$ | domain of sidewalk proportion | $B_{k,t} \subseteq R_{(0,1)}$ |
| $QN_{od,t}$ | vehicular trips from $o$ to $d$ at $t$ | |
| $QP_{od,t}$ | pedestrian trips from $o$ to $d$ at $t$ | |
| $o, d$ | origin point and destination point | $o \in O$ |
| $k$ | edge(agent) | $k \in K$ |
| $t$ | simulation time step | $t \in T$ |
| $h$ | observation step | $h \in H_t$ |
| $n$ | vehicle | $n \in N_{k,h}$ |
| $p$ | pedestrian | $p \in P_{k,h}$ |
| $v_{n,k,h}$ | velocity of vehicle $n$ on $k$ at $h$ | $v_{n,k,h} \in R_{[0,vm]}$ |
| $v_{p,k,h}$ | velocity of pedestrian $p$ on $k$ at $h$ | $v_{p,k,h} \in R_{[0,pm]}$ |
| $f^d_{n,k,h}$ | unit AV flow on $k$ towards $d$ at $h$ | $f^d_{n,k,h} \in N_{\{0,1\}}$ |
| $f^d_{p,k,h}$ | unit pedestrian on $k$ towards $d$ at $h$ | $f^d_{p,k,h} \in N_{\{0,1\}}$ |
| $w_k$ | width of the edge $k$ | $w_k \in R_{(0,+\infty)}$ |
| $\psi_k$ | facility belt proportion of $k$ | $\psi_k \in R_{(0,1)}$ |
| $l_{k,t}$ | number of lanes on edge $k$ at $t$ | $l_{k,t} \in L_{k,t}$ |
| $\beta_{k,t}$ | sidewalk proportion of $k$ at $t$ | $\beta_{k,t} \in B_{k,t}$ |

## III. METHODOLOGY

### A. Problem Definition

A multi-agent system is proposed to represent a road network that consists of multiple edges. Each edge $k \in K$ may be different from others in length, width ($w_k$), facility belt proportion ($\psi_k$), sidewalk proportion ($\beta_k$), carriageway proportion ($1-\beta_k-\psi_k$) and the number of driving lanes ($l_k$). Fig.1 demonstrates the ROW composition of an edge of a network. The width of the street-side can be expressed as $(\psi_k+\beta_k)w_k$, while that of the carriageway equals $(1-\psi_k-\beta_k)w_k$.

A discrete-time transport system evolves ROW compositions of all edges at each step $t \in T$, where $T$ denotes a set comprising a day's worth of time steps. Decision variables include the sidewalk proportion $\beta_{k,t} \in B_{k,t}$ and the number of driving lanes $l_{k,t} \in L_{k,t}$, where $B_{k,t}$ and $L_{k,t}$ are their respective legal domains at $t$. While, $w_k$ and $\psi_k$ are considered as constants.

An edge only acknowledges its local traffic states, namely the number of AVs $|N_{k,h}|$ and pedestrians $|P_{k,h}|$ at each observation $h \in H_t$, where $H_t$ enumerates all observations at a step. Table. II tabulates the parameters for modelling this transport system.

In this paper, we present two distinct learning paradigms for instructing ROW decisions. In the first place, a centralised learning paradigm is devised which considers all edges to co-evolve through a shared deterministic policy function. Even though any local changes to respective ROW composition may affect the future global traffic states to varied extent, their contributions are indiscriminately rewarded. Eq. 2 expresses the expected total reward $r_{k,t}$ per edge under this centralised learning policy, which accumulates the immediate reward $g_{k,t}$ in a discounted norm.

In contrast, the distributive learning policy requires an edge to manage its local evolving patterns, i.e., they collect their learning experiences, update policy functions, and obtain rewards independently, as per Eq. 3.

$$\max_{\beta_{k,t}\in B_{k,t}, l_{k,t}\in L_{k,t}} r_{k,t} = \mathop{E}_{k \sim K} \sum_{t=t'}^{T} \gamma^{t-t'} g_{k,t} \quad (2)$$

$$\max_{\beta_{k,t}\in B_{k,t}, l_{k,t}\in L_{k,t}} r_{k,t} = \sum_{t=t'}^{T} \gamma^{t-t'} g_{k,t} \quad (3)$$

$$g_{k,t} = g^{veh}_{k,t} + g^{ped}_{k,t} + g^{act}_{k,t} \quad (4)$$

$$g^{veh}_{k,t} = \frac{1}{|H_t| \cdot |N_{k,t}|} \sum_{h=0}^{H_t} \sum_{n=0}^{N_{k,t}} \frac{v_{n,k,h}}{vm} (1-\beta_{k,t}, l_{k,t}) \quad (5)$$

$$g^{ped}_{k,t} = \frac{1}{|H_t| \cdot |P_{k,t}|} \sum_{h=0}^{H_t} \sum_{p=0}^{P_{k,t}} \frac{v_{p,k,h}}{pm} (\beta_{k,t}, l_{k,t}) \quad (6)$$

$$g^{act}_{k,t} = \beta_{k,t} + \psi_k \quad (7)$$

Eq. 4 explains that the immediate reward $g_{k,t}$ sums up quadratic gains estimated from traffic flow efficiencies of AVs $g^{veh}_{k,t} \in R_{[0,1]}$, pedestrian movements $g^{ped}_{k,t} \in R_{[0,1]}$, and the street-side proportion, denoted as $g^{act}_{k,t} \in R_{(0,1)}$. All three quadratic terms are weighted by 1. Eq. 5 estimates the ratio of an AV's driving speed $v_{n,k,h}$ to the maximum velocity $vm$. Similarly, Eq. 6 calculates that ratio between walking speed $v_{p,k,h}$ and the maximum speed $pm$ of a pedestrian. Finally, Eq. 7 indicates that the street-side proportion is equal to the sum of the sidewalk proportion $\beta_{k,t}$ and facility belt proportion $\psi_k$.

### B. Reinforcement Learning Model

We outline the key elements of the Multi-Agent Markov Decision Process (MAMDP) of our RL model as follows:

- K  set of multi-agents (edges) $k \in K$.
- $S_t$  global state of the system. $S_t = \{s_{k,t}\} \subset R^{|K|}$ and $s_{k,t}$ denotes partial states observed by local agents.
- $A_t$  collective actions taken by multi-agents, $A_t = \{a_{k,t}\} \subset R^{|K|}$, and $a_{k,t}$ equals the decision variable $\beta_{k,t}$.
- $R_t$  joint cumulative future rewards of all agents and $R_t = \{\phi r_{k,t}\} \subset R^{|K|}$, where $\phi = 1,000$ represents an amplifier for the convenience of numerical analysis.
- $\mu$  a deterministic policy that maps a given state to an action, which is parameterised by a neural network $\theta^\mu$.

The workflow of our model is illustrated in Fig. 2. This schematic comprises a microscopic traffic simulation phase and an off-policy learning phase. First, at step $t$, the model configures the road network, the physical dynamics of users and a synthesised travel plans. Then it outputs traffic states ($S_t$) and rewards ($R_t$). Note that the travel plan could also be extracted from live image streams captured by traffic cameras. The model determines decision variables $\beta_{k,t}$ and $l_{k,t}$ as per its policy function, an additive noise and a greedy algorithm. In the second phase, our model carries out an off-policy learning using sampled experience.

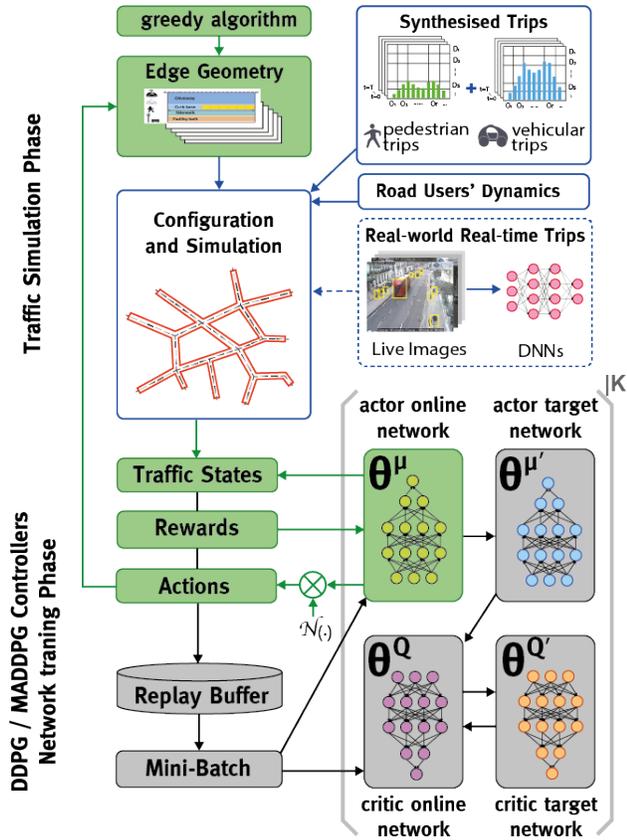

Fig. 2. Framework of microscopic traffic simulation-based Reinforcement Learning model

*1) Actions*: Three steps are involved in determining the decision variables $\beta_{k,t+1}$ and $l_{k,t+1}$ for a new step. First, road edges make control decisions $a_{k,t+1}$ relying on the deterministic policy functions and a decaying noise $N(\cdot)$, as expressed in Eq. 8. As discussed in [49], this noise disturbance follows the decaying-$\epsilon$-greedy policy, which initially encourages exploration in the action space, and generally converges to the mean, i.e. zero in this study.

As displayed in Eq. 9, The degree of randomness of exploration is limited in part by the deviation of the distribution ($\sigma$) and a decaying coefficient $\epsilon \in R_{(0,1)} \cap \epsilon \gg 0$. It is estimated that 99.73% of generated noise values would fall between the range of $R_{[-\sigma,\sigma]}$ at the $i^{th}$ epoch. This approach strikes a balance between exploration and exploitation within a finite learning horizon.

$$a_{k,t+1} = \mu(s_t|\theta^\mu_t) + N(\sigma_i) \quad (8)$$

$$\sigma_{i+1} = \epsilon \cdot \sigma_i \quad (9)$$

$a_{k,t+1}$ is then clipped against a trust region following Eq. 10. The lower bound regulates the minimum width of a sidewalk (1.5m). Meanwhile, the upper bound considers at least one 3.5m-wide lane should be provided for emergency.

$$a_{k,t+1} = \mathrm{clip}\left(a_{k,t+1}, \frac{1.5}{w_k}, \frac{w_k - \psi_k - 3.5}{w_k}\right) \quad (10)$$

Eq. 11 formulates a greedy algorithm to approximate the optimal number of driving lanes ($l^*_{k,t+1}$). To determine the number of lanes, the estimated edge width is minimised in comparison to the actual edge width. Afterwards, the model determines the decision variables for performing the sidewalk proportion ($\beta_{k,t+1}$), as per Eq. 12.

$$l^*_{k,t+1} = \min_{l_{k,t+1}} \lceil (1 - a_{k,t+1} - \psi_k)w_k - 3.5 l_{k,t+1} \rceil \quad (11)$$

$$a_{k,t+1} = \beta_{k,t+1} = 1 - \psi_k - \frac{3.5 l_{k,t+1} - 0.5}{w_k} \quad (12)$$

*2) Observation and Transition of Traffic States*: In this paper, we establish a discrete-time stochastic AI Gym environment [50] that agents will interact with using the open-source Simulation of Urban MObility (SUMO) software [51]. Through SUMO's Traffic Control Interface (TraCI), edge-level traffic states are continuously retrieved and the ROW compositions of the road network are configured iteratively.

The local traffic state $s_{k,t}$ is computed as the expectation of all observed states $s_{k,h}$ following Eqs. 13 and 14. $q \in N_{\{0,1\}}$ distinguishes the class of road users, namely $q=0$ represents pedestrian flow and $q=1$ denotes AV traffic flow. Eq. 15 accumulates flows $e^q_{k,h} \in N_{[0,+\infty)}$ by respective travel modes, where $f^d_{n,k,h} \in N_{\{0,1\}}$ and $f^d_{p,k,h} \in N_{\{0,1\}}$ represent an operational and an active pedestrian within scope.

$$s_{k,t} = \frac{1}{|H_t|} \sum_{h=0}^{|H_t|} s_{k,h} \quad (13)$$

$$s_{k,h} = \{e^q_{k,h} | q \in N_{[0,1]}\} \quad (14)$$

$$e^q_{k,h} = q \sum_{d=0}^{|D:N_{k,t}|} \sum_{n=0}^{|D:P_{k,t}|} f^d_{n,k,h} + (1-q) \sum_{d=0}^{|D:P_{k,t}|} \sum_{p=0}^{|D:P_{k,t}|} f^d_{p,k,h} \quad (15)$$

If possible, all scheduled trips listed in $QN_{od,t}$ and $QP_{od,t}$ should arrive at respective destination $d$ before the last observation step $h = |H_t|$. However, if a trip is still in operation, or has yet to depart, it is re-assigned to the next step, as per Eqs. 16 and 17. Note that $|QN_{od,t+1}|$ and $|QP_{od,t+1}|$ express the cardinalities of respective scheduled travel plans.

$$|QN_{od,t+1}| = |QN_{od,t+1}| + \sum_{k=0}^{H_t} f^d_{n,k,h=|H_t|} \quad (16)$$

$$|QP_{od,t+1}| = |QP_{od,t+1}| + \sum_{k=0}^{H_t} f^d_{p,k,h=|H_t|} \quad (17)$$

*3) Rewards:* The reward $r_{k,t}$ estimates the expectation of future gains upon a sequence of ROW decisions, recalling Eqs. 2-7. The immediate reward is derived directly as feedback from agents' interactions with the system from simulation. While it has a quadratic form to accumulate all obtained gains, it differs fundamentally from those explicit reward functions of model-based RLs or COCPs.

By multiplying $\phi=1,000$ times, the reward ceiling becomes 3,000. Let I comprise all designated training epochs $i \in I$. If all epochs begin training the model at time step zero uniformly, denoted as $e=0$, it would result in the same initial state and possibly similar subsequent transition patterns. Consequently, exploration in state space may be insufficient and training may be biased [52]. Therefore, following Eq. 18, for the first 20 epochs, the model begins learning at $e=0$, and for the rest, it generates a random starting slot apart from last two slots. Accordingly, the edge reward ($r_i$) of epoch $i \in I$ is calculated as per Eq. 19.

$$e = \begin{cases} 0, & \text{if } i \in N_{[0:19]} \\ rnd(1, |T| - 2), & \text{if } i \in N_{[20:|I|-1]} \end{cases} \quad (18)$$

$$r_i = \frac{1}{(|T| - e) \cdot |K|} \sum_{k=0}^{K} \sum_{t=0}^{T} r_{k,t,i} \quad (19)$$

### C. Architecture and Learning Mechanism of DDPG

In this article, the Deep Deterministic Policy Gradient (DDPG) algorithm [20] is presented as a method for selecting the optimal ROW plans. We deploy a centralised AC neural network structure and a global experience replay buffer. The AC architecture consists of an actor online network ($\mu$), an actor target network ($\mu'$), a critic online network ($Q$) and a critic target network ($Q'$), of which their neural weights are denoted by $\theta$. Moreover, the buffer (D) indiscriminately stores transition tuples and randomises the sampling of data for off-policy learning. Let a mini-batch M contains the sampled transition tuples denoted as $T_m = \langle s_m, a_m, r_m, s_{m+1} \rangle$, where $m \in M$.

Through back propagation, gradient descent, and loss regression, the critic online network is updated [20]. Let $y_m$ denotes an approximator of the Q-value $Q(s_m, a_m|\theta^Q)$. The piecewise Eq. 20 elaborates that if the sampled data represents the last time step, then $y_m$ equals the state value of the final step; Otherwise, it equals the sum of the accumulative discounted Q-value. Then, we calculate the critic loss ($L_Q$) of approximated Q-value and the authentic Q-value using the Huber's loss function ($\delta=1.0$). This loss function, as demonstrated in piecewise Eqs. 21 and 22, outperforms the Mean-Squared Error (MSE) loss function, in that it is less sensitive to outliers [53].

$$y_m = \begin{cases} r_m, & \text{if } s_m = s_t \cap t = T \\ r_m + \gamma Q[s_{m+1}, \mu(s_{m+1}|\theta^{\mu'})|\theta^{Q'}], & \text{otherwise} \end{cases} \quad (20)$$

$$\xi_m = y_m - Q(s_m, a_m|\theta^Q) \quad (21)$$

$$L_{Q,\delta} = \begin{cases} \mathbb{E}_{m \sim M} \frac{1}{2}\xi_m^2, & \text{if } |\xi_m| \leq \delta \\ \delta \mathbb{E}_{m \sim M} |\xi_m| - \frac{1}{2}\delta, & \text{otherwise} \end{cases} \quad (22)$$

Based on previous Eqs. 8 and 9, actor online network determines the next action. Applying the learnt Q function, $\mu$ estimates its policy function by maximising the expected return. The actor loss, denoted by $J_\mu$, is estimated following Eq. 23. Furthermore, we calculate the derivative of the expectation of gradients on both critic online network $Q$ and the actor online network $\mu$, as per Eq. 24.

$$J_\mu = \mathbb{E}_{m \sim M} [Q(s_m, a_m|\theta^\mu)] \quad (23)$$

$$\nabla_{\theta^\mu} J_\mu \approx \mathbb{E}_{m \sim M} [\nabla_a Q(s_m, a_m|\theta^Q) \nabla_{\theta^\mu} \mu(s_m|\theta^\mu)] \quad (24)$$

Target AC networks are initialised and updated based on their respective online counterparts. The mechanism of updating uses a delayed copying strategy that approximates the target parameters at each step. Concretely speaking, following Eqs. 25 and 26, the new weights inherit a portion $1 - \eta$ of their current weights while adding a portion $\eta$ from their online counterparts, where $0 < \eta \ll 1$ denotes a copying coefficient.

$$\theta^{Q'} \leftarrow (1 - \eta)\theta^{Q'} + \eta\theta^Q \quad (25)$$

$$\theta^{\mu'} \leftarrow (1 - \eta)\theta^{\mu'} + \eta\theta^\mu \quad (26)$$

Algorithm.1 presents the workflow of this DDPG algorithm. It initially inputs a series of hyperparameters and travel schedules. At step $t$ of epoch $i$, all agents select respective actions $a_{k,t,i}$ and determine the numbers of driving lanes $l_{k,t,i}$ as per Eqs. 8-12. The model re-configures the ROW compositions of all edges and interacts with the new road layout via SUMO TraCI for the return of a joint state $S_{t+1,i}$ and discrete rewards $r_{k,t,i}$, which is redistributed as per Eq. 2. Then, the model stores a collection of transition tuples to a global buffer D and performs the centralised off-policy learning following Eqs. 23-26.

### D. Architecture and Learning Mechanism of MADDPG

Aside from DDPG, we also propose a Multi-Agent DDPG (MADDPG) algorithm to assess the performance of adopting the distributed learning strategy. For MADDPG, each edge can learn from its specific context rather than knowledge that is publicly shared. The objectives are independently optimised

**Algorithm 1** Pseudocode of DDPG Algorithm

**Input** I, $c$, M, T, K, S$_0$, $\eta$
**Initialise** $\theta^Q, \theta^\mu, \theta^{Q'} \leftarrow \theta^Q, \theta^{\mu'} \leftarrow \theta^\mu$, D $\leftarrow c$, T $\leftarrow$ |M|
**for** $i = 0 : (|I| - 1)$ **do**:
  Initialise $e$ see Eq. 18

  % ON-POLICY ACTING AND SIMULATION
  **for** $t = e : (|T| - 1)$ **do**:
    **for** $k = 0 : (|K| - 1)$ **do**:
      A$_{t,i} \leftarrow a_{k,t,i}$, see Eq. 8-12
    S$_{t+1,i}, r_{k,t,i} \leftarrow$ TraCI(A$_{t,i}, l_{k,t,i}$) see Eq. 2
    **for** $k = 0 : (|K| - 1)$ **do**:
      $\{s_{k,t+1,i}\} \leftarrow$ S$_{t+1,i}$
      D $:= \langle a_{k,t,i}, s_{k,t,i}, r_{k,t,i}, s_{k,t+1,i} \rangle$

  % OFF-LINE LEARNING
  **random sample** T $\leftarrow$ D, |M|
  **for** $m = 0 : (|M| - 1)$ **do**:
    $y_m \leftarrow \theta^{Q'}$, T$_m$, see Eq. 20
    $\theta^Q, \theta^\mu, \theta^{Q'}, \theta^{\mu'} \leftarrow y_m, \theta^Q, \theta^\mu, \eta$, see Eq. 21-26
**Return** $\theta^Q, \theta^\mu, \theta^{Q'}, \theta^{\mu'}$

following Eq.3. The MADDPG architecture includes multiple AC networks and independent replay buffers D$_k$. Algorithm.2 demonstrates its learning procedure. Further, as per previous studies [18], [20], the notions and values tuned for hyperparameters are presented in Table. III.

**Algorithm 2** Pseudocode of MADDPG Algorithm

**Input** I, $c$, M, T, K, S$_0$, $\eta$
**Initialise** $\theta^Q_k, \theta^\mu_k, \theta^{Q'}_k \leftarrow \theta^Q_k, \theta^{\mu'}_k \leftarrow \theta^\mu_k$, D$_k \leftarrow c$, T$_k \leftarrow$ |M|
**for** $i = 0 : (|I| - 1)$ **do**:
  Initialise $e$ see Eq. 18

  % ON-POLICY ACTING AND SIMULATION
  **for** $t = e : (|T| - 1)$ **do**:
    **for** $k = 0 : (|K| - 1)$ **do**:
      A$_{t,i} \leftarrow a_{k,t,i}$, see Eqs. 8-12
    S$_{t+1,i}, r_{k,t,i} \leftarrow$ TraCI(A$_{t,i}, l_{k,t,i}$) see Eq. 3
    **for** $k = 0 : (|K| - 1)$ **do**:
      $\{s_{k,t+1,i}\} \leftarrow$ S$_{t+1,i}$
      D$_k := \langle a_{k,t,i}, s_{k,t,i}, r_{k,t,i}, s_{k,t+1,i} \rangle$

  % OFF-LINE LEARNING
  **for** $k = 0 : (|K| - 1)$ **do**:
    **random sample** T$_k \leftarrow$ D$_k$, |M|
    **for** $m = 0 : (|M| - 1)$ **do**:
      $y_{k,m} \leftarrow \theta^{Q'}_k$, T$_m$, see Eq. 20
      $\theta^Q_k, \theta^\mu_k, \theta^{Q'}_k, \theta^{\mu'}_k \leftarrow y_{k,m}, \theta^Q_k, \theta^\mu_k, \eta$, see Eqs. 21-26
**Return** $\theta^Q_k, \theta^\mu_k, \theta^{Q'}_k, \theta^{\mu'}_k$

TABLE III
NOTIONS AND HYPERPARAMETERS FOR DDPG AND MADDPG ALGORITHMS

| Notations | Specifications | Domains |
|---|---|---|
| I | set of training epochs | |I|=150 |
| M | mini-batch for training | |M|=64 |
| $i$ | index of training epochs | $i \in$ I |
| $m$ | index of training samples | $m \in$ M |
| $t$ | index of time slots | $t \in$ T |
| $e$ | index of starting slots | $e \in N_{[1:|T|-2]}$ |
| $c$ | replay buffer capacity | 100,000 |
| $\phi$ | rewards amplifier | 1,000 |
| $\eta$ | delayed copying coefficient | 0.005 |
| $\gamma$ | discount factor | 0.99 |
| $\epsilon$ | noise decaying factor | 0.99 |
| $\sigma$ | noise standard deviation | 0.20, 0.40, 0.60 |

IV. SIMULATION SET-UP

We used a Mac OS v12.01 computer to train the model. It is equipped with a processor of 2.6GHz 6-Core Intel Core i7-9750H, 16GB of RAM and AMD Radeon Pro 5300M GPU.

*A. Road Networks Configuration*

We designed a parametric generation function in SUMO to create customised road network samples. Among the samples are simple road components such as T-junctions, intersections, and roundabouts. Moreover, it can be integrated with OpenStreetMap (OSM) data to configure real-world road networks.

This study utilised four road geometries and a real-world road network for testing. As illustrated in Fig. 3, testing cases include a street section, a T-junction, an intersection and a roundabout. The street section consists of two counter-directional edges (|K|=2) in a length of 100$m$. The initial layout has three lanes in each direction and each 3.5$m$ wide. The total edge width is $w$=13$m$, and the facility belt ratio equals $\psi = \frac{1.5m}{13m}$. Following such a basic setting, the remaining three cases are developed with respective dimensions of |K|=6, 8, and 12. The straight edges of the roundabout are 50$m$ long.

Next, we chose the Belgrave road network as a real-world case for study, which is demonstrated in Fig. 4. Located in London's busiest and central districts (Latitude: 51.497020, Longitude: −0.151790), this complex network includes 326 edges (|K|=326), 58 T-junctions, 23 intersections, one radial intersection, and three irregular roundabouts. Edges are topologically heterogeneous regarding lengths, ROW compositions and connections. On average, the total edge width is 8.50$m$, the length is 49.30$m$, and the street-side width is 2.28$m$.

*B. Dynamics of Road Users and Travel Demand*

We calibrated driving behaviours of AVs and pedestrian movement dynamics based on previous studies [18], [54]. Namely, AVs can travel at a top speed $v_m$=30$km/h$, the time headway is 0.60$s$. The maximum walking speed $p_m$ of pedestrians is 1.3$m/s$. Each OD pair travel demand is planned on a half-hour schedule, contributing to |T|=48 time steps. The flow rates of AVs and pedestrians demonstrate bimodal distribution [55], which equals 114$veh/h$ and 21$p/h$ on average per *o-d* pair.

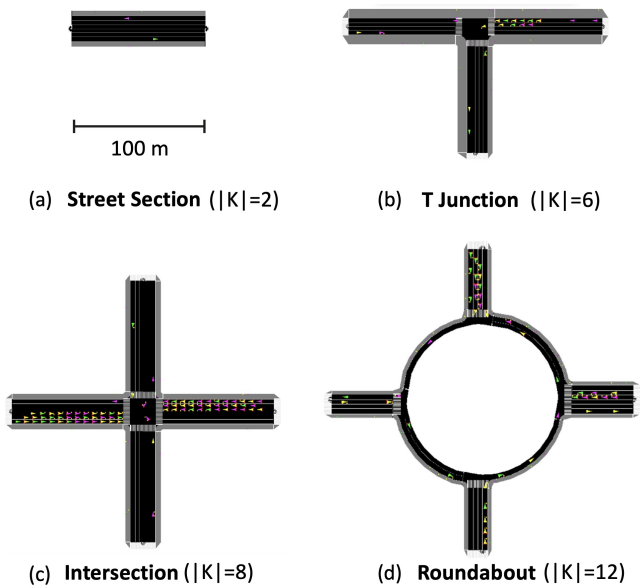

Fig. 3. Basic road components for testing. (a) Street section case, (b) T-junction case, (c) Intersection case, (d) Roundabout case

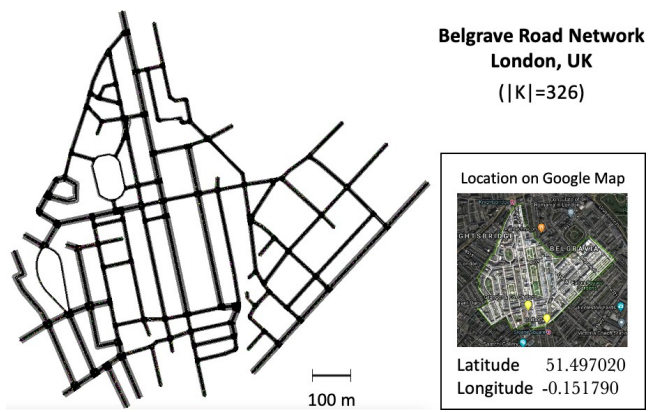

Fig. 4. The configuration and surroundings of the London Belgrave Network

## V. RESULTS AND DISCUSSION

DDPG and MADDPG algorithms were applied separately to each case for training in 150 epochs (avg. 4080 time steps). For these five cases, DDPG takes 110, 248, 513, 388, and 848*mins* to compute. For MADDPG, those are 71, 108, 184, 195 and 503*mins*, respectively. In comparison with DDPG, MADDPG's total running time is 49.64% shorter, demonstrating its greater algorithmic efficiency.

### A. Results on Road Component Cases

Fig. 5 presents the epoch-wise rewards under four road geometries, namely the street section, the T-junction, the intersection and the roundabout case. X-axes index training epochs and Y-axes on either side mark the ranges of rewards of DDPG (———) and MADDPG (———).

In independent courses of training, DDPG and MADDPG improve rewards by 6.30% and 4.62%, respectively. MADDPG outperforms DDPG in benchmark rewards, optimal

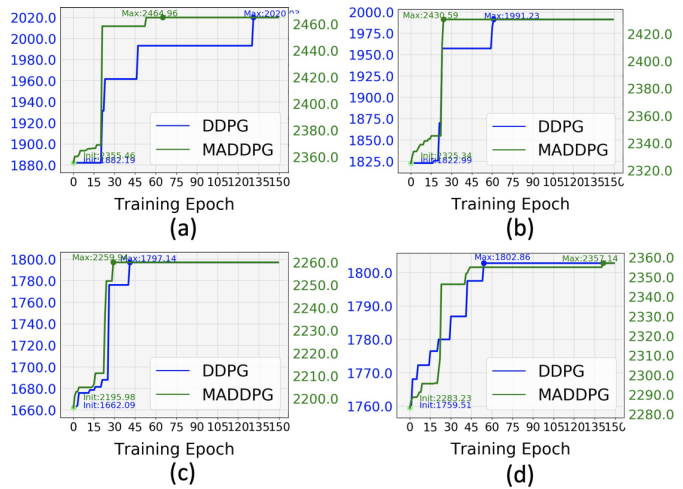

Fig. 5. Convergence patterns of DDPG (———) and MADDPG (———) under cases of four road components. (a) Street section case, (b) T-junction case, (c) Intersection case, (d) Roundabout case.

rewards, and sheer improvements by 42.56%, 24.98%, and 33.62%, respectively.

As the dimension increases, both the benchmark rewards and the learnt optima decrease. In the roundabout case, for example, the initial rewards are -6.52% (DDPG) and -3.16% (MADDPG) lower than those in the street section case, while the found optima decreased by -10.75% and -8.10%, respectively.

In cases of the street section, T-junction, and intersection, MADDPG reaches the convergence zone earlier than DDPG, whereas DDPG is faster only in the case of the roundabout, but with 23.51% smaller in the optima. MADDPG and DDPG exhibit divergent convergence curvature, as the former shows a steady and constant convergence patterns, whereas the latter stagnates at a few local optima.

### B. Results on the Real-world Road Network

Simulation results of the real-world case show an average AV and pedestrian throughput of 1,664*veh/h* and 1,289*p/h*, respectively. Driving and walking speeds have been improved to some extent. Namely, Table. IV compares the changes in observed speeds in the first and last 30 training epochs.

TABLE IV
VARIATIONS IN THE AVERAGE WALKING AND DRIVING SPEEDS

|        | Speed          | Ep.0-30 | Ep.120-149 | Changes |
|--------|----------------|---------|------------|---------|
| MADDPG | Driving (*km/h*) | 20.401  | 20.772     | +15.44% |
|        | Walking (*m/s*)  | 0.848   | 0.865      | +2.00%  |
| DDPG   | Driving (*km/h*) | 18.065  | 18.076     | +0.06%  |
|        | Walking (*m/s*)  | 0.761   | 0.762      | +0.13%  |

The rewards and action values are displayed in Fig. 6. Fig. 6(a) shows their reward patterns. The benchmark reward of MADDPG is greater than those of DDPG by 65.51%. Similarly, the optima found by MADDPG is higher than DDPG by 505.79. Besides, as |K| substantially increased to 326, the optima explored by DDPG and MADDPG sharply

decrease by 61.75% and 48.12%, compared with the street section case.

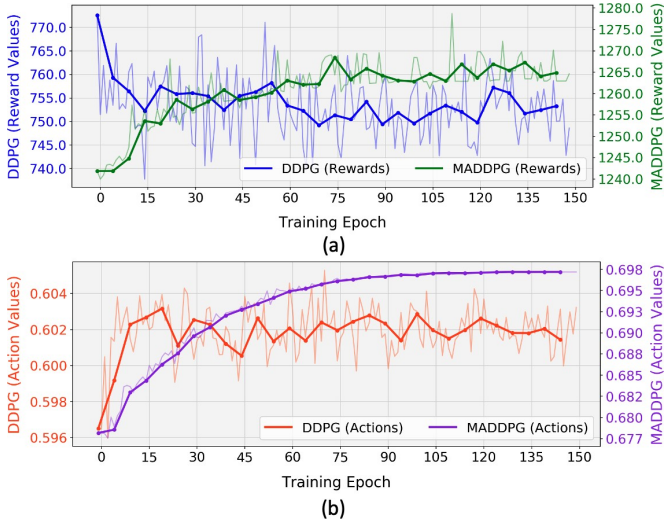

Fig. 6. Rewards and actions values for the Belgrave case. (a) Rewards obtained by DDPG (——) and MADDPG (——). (b) Actions taken by DDPG (——) and MADDPG (——).

Fig. 6(b) demonstrates the evolving patterns of action values for DDPG and MADDPG. Both the initial and optimal values of MADDPG are superior to its counterpart. At the initial epoch, the action values are 0.596 and 0.678, respectively. With a favourable benchmark, MADDPG persistently optimises its ROW plans and steadily increases the proportion of street-side from 0.678 to 0.697. However, DDPG explores an optima of 0.603 and increases the street-side proportion by 1.34%, which is underperformed compared with MADDPG.

MADDPG demonstrates stronger signs of convergence guarantee, with a monotonic increasing trend in the action curve, whose optima converges around the $120^{th}$ epoch. Afterwards, the pattern shows a gradually narrowing amplitude. In contrast, the patterns of action values of DDPG seemly show steep learning curve at the initial phase, while signalling a slightly weaker convergence guarantee later on, with constant vibrations in a range of $R_{[0.600, 0.604]}$.

In summary, tested against the real-world network case, the DDPG and MADDPG have separately optimised their control strategies and improved their respective reward to 55.90% and 72.61% of the ceiling (3,000). Notably, MADDPG outperforms DDPG regarding benchmark rewards (25.35%), quality of optima (24.58%), optimal action values (13.49%), convergence speed and persistence of optimisation. These findings provide evidence that distributed learning is more effective in controlling heterogeneous agents than centralised learning. Such performance gaps widened when comparing outcomes of previous four cases with homogeneous agents.

While DDPG performed poorly than MADDPG, it provided the essential architecture and a baseline for its multi-agent counterpart. In spite of its poor parallelism and scalability, DDPG is theoretically more sample efficient than MADDPG. It means that, even when a number of edges could not receive adequate observations (e.g. extremely low traffic flow per edge), DDPG may still be able to collect sufficient samples for centralised training. In such circumstances, the MADDPG controller would be invalid. Additionally, we are interested in finding out whether adopting a 'sparse reward' strategy would improve its convergence performance and the quality of optima.

### C. Impact of Variations in Action Exploration

Variation of noise deviation significantly affects learning performance. We have tested three scenarios with different noise standard deviation, as per Figs. 7(a)-(c), namely $\sigma$= 0.2, 0.4 and 0.6, whereas the rest variables remain constant.

Intuitively, all three patterns have shown convergence trends following the decaying-$\epsilon$-greedy policy, with convergence rates of ±0.0005/$ep$, ±0.0012/$ep$ and ±0.0018/$ep$ towards zero. However, their initial distributions differ significantly, with $\sigma$=0.6 having the broadest exploration veering off the mean.

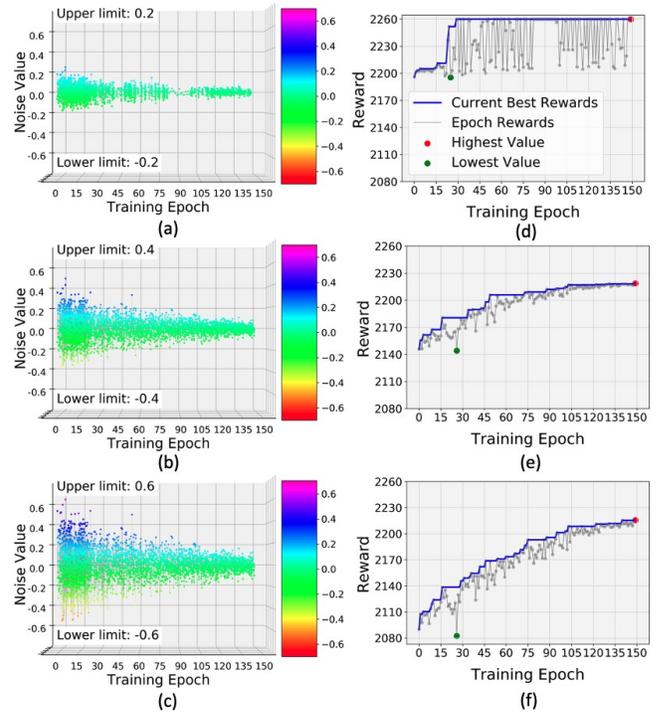

Fig. 7. Distributions of noise and corresponding learning curves under different noise scenarios. Distributions of disturbance: (a) $\sigma$=0.2, (b) $\sigma$=0.4, (c) $\sigma$=0.6. Learning curves: (d) $\sigma$=0.2, (e) $\sigma$=0.4, (f) $\sigma$=0.6.

Figs. 7(d)-(f) demonstrate the learning performances of corresponding noise conditions. The benchmark reward of $\sigma$=0.20 is higher than $\sigma$=0.40 and 0.60 by 2.33% and 5.15%, and with higher (1.83% and 1.96%) explored optima.

With $\sigma$=0.20, the early rewards obtained before $5^{th}$ epoch have already surpassed 2,200. It quickly converges to its optima at the $30^{th}$ epoch. The improvement from 2,144.26 to 2,200 takes $\sigma$=0.40 an additional 40 epochs. Likewise, for $\sigma$=0.60, its convergence speed is even slower. At the end of the learning process, it would take another 90 epochs to reach a stable convergence of 2,082.39 to 2,200.

We conclude that a lower noise standard deviation may lead to favourable benchmark rewards, quicker convergence rates, and better optima in this case. Even though greater noise disturbance allows broader initial exploration, it does not guarantee the convergence given the finite horizon instances.

## VI. CONCLUSION

In this study, we designed an RL model to evolve ROW configurations to accommodate varying patterns of AV traffic flows and pedestrian movements. The configuration process was facilitated by a microscopic traffic simulator, which functions as an AI Gym environment. A balance was pursued between traffic flow efficiency and the space allocated to non-vehicular road users. We further employed two different solvers - DDPG and MADDPG - for the representation of the centralised and distributed learning strategies, respectively.

Experimental results showed that DDPG and MADDPG optimised ROW compositions, which are active mobility-friendly, and improved traffic flow efficiency. MADDPG outperformed DDPG in computational time, obtained benchmark rewards, best cumulative rewards, optimal action values and convergence speed. In addition, the learning patterns of MADDPG demonstrated a stronger convergence guarantee than its counterpart. Moreover, the distributive learning strategy outperformed its centralised counterpart in heterogeneous multi-agent systems.

There is potential to improve this study regarding the demands of multiple road space stakeholders. Second, we plan to improve the RL method regarding sample efficiency and exploration policy. In addition, new RL algorithms would be tested for better solutions regarding the overestimation of Q-values and the hyperparameters sensitivity problem. To better motivate the reward mechanism, the current reward architecture might be re-designed. For instance, the quadratic terms of the immediate reward could be weighted differently depending on the time of day or location.

We would apply more realistic road configurations and travel demand patterns for testing. Meanwhile, this RL method is promising to be integrated with intelligent road infrastructure for an end-to-end allocation of ROWs.

## REFERENCES

[1] W. Zhang, S. Guhathakurta, and E. B. Khalil, "The impact of private autonomous vehicles on vehicle ownership and unoccupied vmt generation," *Transportation Research Part C: Emerging Technologies*, vol. 90, pp. 156–165, 2018.
[2] S. International, "Sae levels of driving automation refined for clarity and international audience," 2021, [Accessed 19-Jan-2022]. [Online]. Available: https://www.sae.org/blog/sae-j3016-update
[3] D. Omeiza, H. Webb, M. Jirotka, and L. Kunze, "Explanations in autonomous driving: A survey," *IEEE Transactions on Intelligent Transportation Systems*, 2021.
[4] J. Van Brummelen, M. O'Brien, D. Gruyer, and H. Najjaran, "Autonomous vehicle perception: The technology of today and tomorrow," *Transportation research part C: emerging technologies*, vol. 89, pp. 384–406, 2018.
[5] R. Utriainen, "The potential impacts of automated vehicles on pedestrian safety in a four-season country," *Journal of Intelligent Transportation Systems*, vol. 25, no. 2, pp. 188–196, 2020.
[6] ERTRAC, "Automated driving roadmap," 2017, [Accessed 19-Apr-2022]. [Online]. Available: https://www.ertrac.org/uploads/images/ERTRAC_Automated_Driving_2017.pdf
[7] T. Litman, *Autonomous vehicle implementation predictions*. Victoria Transport Policy Institute Victoria, Canada, 2017.
[8] R. Snyder, "Street design implications of autonomous vehicles," 2018, [Accessed 11-May-2022]. [Online]. Available: https://www.cnu.org/publicsquare/2018/03/12/street-design-implications-autonomous-vehicles
[9] S. E. Shladover, D. Su, and X.-Y. Lu, "Impacts of cooperative adaptive cruise control on freeway traffic flow," *Transportation Research Record*, vol. 2324, no. 1, pp. 63–70, 2012.
[10] J. Moavenzadeh and N. S. Lang, "Reshaping urban mobility with autonomous vehicles: Lessons from the city of boston," in *World Economic Forum*, 2018.
[11] A. R. Alozi and K. Hamad, "Quantifying impacts of connected and autonomous vehicles on traffic operation using micro-simulation in dubai, uae." in *VEHITS*, 2019, pp. 528–535.
[12] Department of Transport, Government of UK, "The highway code, hierarchy of road users," 2022, [Accessed 30-Apr-2022]. [Online]. Available: https://www.gov.uk/guidance/the-highway-code/introduction
[13] D. L. Prytherch, "Legal geographies—codifying the right-of-way: Statutory geographies of urban mobility and the street," *Urban Geography*, vol. 33, no. 2, pp. 295–314, 2012.
[14] NACTO, "Blueprint for autonomous urbanism," 2017, [Accessed 31-Aug-2021]. [Online]. Available: https://nacto.org/publication/bau2
[15] A. Howell, N. Larco, R. Lewis, and B. Steckler, "New mobility in the right-of-way," Urbanism Next - University of Oregon, Tech. Rep. 73, 3 2019.
[16] W. Riggs, B. Appleyard, and M. Johnson, "A design framework for livable streets in the era of autonomous vehicles," *Urban, Planning and Transport Research*, vol. 8, no. 1, pp. 125–137, 2020.
[17] Q. Ye, S. M. Stebbins, Y. Feng, E. Candela, M. Stettler, and P. Angeloudis, "Intelligent management of on-street parking provision for the autonomous vehicles era," in *2020 IEEE 23rd International Conference on Intelligent Transportation Systems (ITSC)*. IEEE, 2020, pp. 1–7.
[18] Q. Ye, Y. Feng, E. Candela, J. Escribano Macias, M. Stettler, and P. Angeloudis, "Spatial-temporal flows-adaptive street layout control using reinforcement learning," *Sustainability*, vol. 14, no. 1, p. 107, 2022.
[19] U. Gunarathna, H. Xie, E. Tanin, S. Karunasekara, and R. Borovica-Gajic, "Real-time lane configuration with coordinated reinforcement learning," in *Joint European Conference on Machine Learning and Knowledge Discovery in Databases*. Springer, 2020, pp. 291–307.
[20] T. P. Lillicrap, J. J. Hunt, A. Pritzel, N. Heess, T. Erez, Y. Tassa, D. Silver, and D. Wierstra, "Continuous control with deep reinforcement learning," *arXiv preprint arXiv:1509.02971*, 2015.
[21] K. Arulkumaran, M. P. Deisenroth, M. Brundage, and A. A. Bharath, "Deep reinforcement learning: A brief survey," *IEEE Signal Processing Magazine*, vol. 34, no. 6, pp. 26–38, 2017.
[22] M. Amirgholy, M. Shahabi, and H. O. Gao, "Traffic automation and lane management for communicant, autonomous, and human-driven vehicles," *Transportation research part C: emerging technologies*, vol. 111, pp. 477–495, 2020.
[23] M. W. Levin and A. Khani, "Dynamic transit lanes for connected and autonomous vehicles," *Public Transport*, vol. 10, no. 3, pp. 399–426, 2018.
[24] B. Yan, P. Wu, and L. Xu, "Optimal design of reserved lanes for automated truck considering user equilibrium," in *2021 IEEE International Conference on Networking, Sensing and Control (ICNSC)*, vol. 1. IEEE, 2021, pp. 1–6.
[25] N. Casas, "Deep deterministic policy gradient for urban traffic light control," *arXiv preprint arXiv:1703.09035*, 2017.
[26] W. Genders, "Deep reinforcement learning adaptive traffic signal control," Ph.D. dissertation, McMaster University, Hamilton, ON, Canada, 2018.
[27] K. Gregg and P. Hess, "Complete streets at the municipal level: A review of american municipal complete street policy," *International journal of sustainable transportation*, vol. 13, no. 6, pp. 407–418, 2019.
[28] E. Dumbaugh and M. King, "Engineering livable streets: A thematic review of advancements in urban street design," *Journal of Planning Literature*, vol. 33, no. 4, pp. 451–465, 2018.
[29] N. A. of City Transportation Officials, *Global street design guide*. Island Press, 2016.
[30] Federal Highway Administration, U.S., "Traffic data computation method, pocket guide," 2018, [Accessed 11-May-2022]. [Online]. Available: https://www.fhwa.dot.gov/policyinformation/pubs/pl18027_traffic_data_pocket_guide.pdf
[31] E. Gravelle and S. Martínez, "Distributed dynamic lane reversal and rerouting for traffic delay reduction," *International Journal of Control*, vol. 91, no. 10, pp. 2355–2365, 2018.

[32] Z. H. Khattak, B. L. Smith, M. D. Fontaine, J. Ma, and A. J. Khattak, "Active lane management and control using connected and automated vehicles in a mixed traffic environment," *Transportation Research Part C: Emerging Technologies*, vol. 139, p. 103648, 2022.

[33] Z. Zhang and S. Tang, "Enhancing urban road network by combining route planning and dynamic lane reversal," in *2021 Thirteenth International Conference on Mobile Computing and Ubiquitous Network (ICMU)*. IEEE, 2021, pp. 1–6.

[34] A. Filos, "Reinforcement learning for portfolio management," *arXiv preprint arXiv:1909.09571*, 2019.

[35] D. Ernst, M. Glavic, F. Capitanescu, and L. Wehenkel, "Reinforcement learning versus model predictive control: a comparison on a power system problem," *IEEE Transactions on Systems, Man, and Cybernetics, Part B (Cybernetics)*, vol. 39, no. 2, pp. 517–529, 2008.

[36] R. Bellman, "The theory of dynamic programming," *Bulletin of the American Mathematical Society*, vol. 60, no. 6, pp. 503–515, 1954.

[37] A. Agrawal, S. Barratt, S. Boyd, and B. Stellato, "Learning convex optimization control policies," in *Learning for Dynamics and Control*. PMLR, 2020, pp. 361–373.

[38] T. M. Moerland, J. Broekens, and C. M. Jonker, "Model-based reinforcement learning: A survey," *arXiv preprint arXiv:2006.16712*, 2020.

[39] P. Beuchat, A. Georghiou, and J. Lygeros, "Alleviating tuning sensitivity in approximate dynamic programming," in *2016 European Control Conference (ECC)*. IEEE, 2016, pp. 1616–1622.

[40] Y. Chebotar, K. Hausman, M. Zhang, G. Sukhatme, S. Schaal, and S. Levine, "Combining model-based and model-free updates for trajectory-centric reinforcement learning," in *International conference on machine learning*. PMLR, 2017, pp. 703–711.

[41] D. Bertsekas, *Reinforcement learning and optimal control*. Athena Scientific, 2019.

[42] J. Bao, G. Zhang, Y. Peng, Z. Shao, and A. Song, "Learn multi-step object sorting tasks through deep reinforcement learning," *Robotica*, pp. 1–17, 2022.

[43] R. S. Sutton and A. G. Barto, *Reinforcement learning: An introduction*. MIT press, 2018.

[44] Y. Yu, "Towards sample efficient reinforcement learning." in *IJCAI*, 2018, pp. 5739–5743.

[45] R. Lincoln, S. Galloway, B. Stephen, and G. Burt, "Comparing policy gradient and value function based reinforcement learning methods in simulated electrical power trade," *IEEE Transactions on Power Systems*, vol. 27, no. 1, pp. 373–380, 2011.

[46] N. Hammami and K. K. Nguyen, "On-policy vs. off-policy deep reinforcement learning for resource allocation in open radio access network," in *2022 IEEE Wireless Communications and Networking Conference (WCNC)*. IEEE, 2022, pp. 1461–1466.

[47] L. Engstrom, A. Ilyas, S. Santurkar, D. Tsipras, F. Janoos, L. Rudolph, and A. Madry, "Implementation matters in deep rl: A case study on ppo and trpo," in *International conference on learning representations*, 2019.

[48] X. Xiong, J. Wang, F. Zhang, and K. Li, "Combining deep reinforcement learning and safety based control for autonomous driving," *arXiv preprint arXiv:1612.00147*, 2016.

[49] A. Ostovar, O. Ringdahl, and T. Hellström, "Adaptive image thresholding of yellow peppers for a harvesting robot," *Robotics*, vol. 7, no. 1, p. 11, 2018.

[50] G. Brockman, V. Cheung, L. Pettersson, J. Schneider, J. Schulman, J. Tang, and W. Zaremba, "Openai gym," 2016. [Online]. Available: https://arxiv.org/abs/1606.01540

[51] P. A. Lopez, M. Behrisch, L. Bieker-Walz, J. Erdmann, Y.-P. Flötteröd, R. Hilbrich, L. Lücken, J. Rummel, P. Wagner, and E. Wießner, "Microscopic traffic simulation using sumo," in *The 21st IEEE International Conference on Intelligent Transportation Systems*. IEEE, 2018. [Online]. Available: https://elib.dlr.de/124092/

[52] M. Coggan, "Exploration and exploitation in reinforcement learning," *Research supervised by Prof. Doina Precup, CRA-W DMP Project at McGill University*, 2004.

[53] E. Kharitonov, "Empirical study of matrix factorization methods for collaborative filtering," in *International Conference on Pattern Recognition and Machine Intelligence*. Springer, 2011, pp. 358–363.

[54] A. Raffin, M. Taragna, and M. Giorelli, "Adaptive longitudinal control of an autonomous vehicle with an approximate knowledge of its parameters," in *2017 11th International Workshop on Robot Motion and Control (RoMoCo)*. IEEE, 2017, pp. 1–6.

[55] S. Susilawati, M. A. Taylor, and S. V. Somenahalli, "Distributions of travel time variability on urban roads," *Journal of Advanced Transportation*, vol. 47, no. 8, pp. 720–736, 2013.

## VII. Biography Section

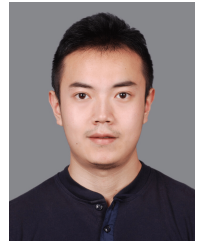

**Qiming Ye** is a PhD student at the Transport Systems and Logistics Laboratory (TSL) in the Department of Civil and Environmental Engineering at Imperial College London. He received the BEng and MEng Degrees in urban planning from Tongji University, with an additional Master's Degree in urban planning and policy design from Politecnico di Milano. His research focuses on smart city evaluation systems and management techniques, i.e. the optimisation and control models for urban space in the era of autonomous vehicles transport.

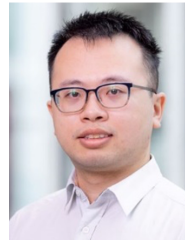

**Yuxiang Feng** is a Research Associate at the Transport Systems and Logistics Laboratory (TSL) in the Department of Civil and Environmental Engineering at Imperial College London. He received a BEng in Mechanical Engineering from Tongji University and an MSc in Mechatronics and PhD in Automotive Engineering from the University of Bath. His main research interests include environment perception, sensor fusion and artificial intelligence for robotics and autonomous vehicles.

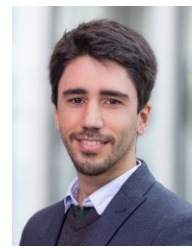

**Jose Javier Escribano Macias** is a Research Associate at the Transport Systems and Logistics Laboratory (TSL), Centre for Transport Studies (CTS) at Imperial College London. He joined CTS in October 2015 as part of the EPSRC Centre for Doctoral Training (CDT) in Sustainable Civil Engineering and was awarded his PhD in March 2021. His research focuses on collaborative vehicle control, optimisation of last-mile logistics, urban air mobility, and machine learning and game theoretical models.

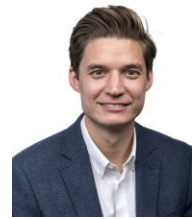

**Marc Stettler** is a Reader in Transport and the Environment in the Department of Civil and Environmental Engineering at Imperial College London. He leads the Transport & Environment Laboratory within the Centre for Transport Studies (CTS) and also the Network of Excellence in Aerosols and Health. He completed his PhD on the impacts of aviation emissions at the University of Cambridge in 2013. His research focuses on evaluating and reducing the effect of transport activity on climate change and air pollution, with particular attention to understanding sources of greenhouse gases (GHGs) and pollutant emissions, especially nanoparticles.

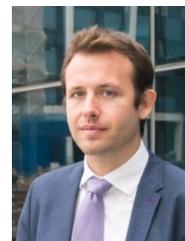

**Panagiotis Angeloudis** is Reader and Head of the Transport Systems and Logistics Laboratory (TSL), based in the Centre for Transport Studies (CTS) at Imperial College London. Before establishing TSL, Panagiotis held a JSPS Research Fellowship at Kyoto University. He previously obtained a PhD in Transportation at Imperial College London and spent periods as a research analyst at DP World and the United Nations in Geneva. His research focuses on the intersection of autonomous systems, multi-agent modelling and network optimisation and their applications to freight distribution and passenger transportation. His research group specialises in developing high-performance, scalable models that capture the interactions between users, providers, infrastructure and operating regimes.